\documentclass[10pt,twocolumn,letterpaper]{article}

\usepackage{wacv}              % To produce the CAMERA-READY version

\usepackage{times}
\usepackage{epsfig}
\usepackage{graphicx}
\usepackage{amsmath}
\usepackage{amssymb}
\usepackage{booktabs}
\usepackage{caption}
\usepackage{subcaption}
\usepackage{multirow}
\usepackage{threeparttable}
\usepackage[export]{adjustbox}
\usepackage{color, colortbl}
\usepackage[none]{hyphenat}
\usepackage[accsupp]{axessibility}

\definecolor{Gray}{gray}{0.9}

\usepackage[pagebackref,breaklinks,colorlinks]{hyperref}

\usepackage[capitalize]{cleveref}
\crefname{section}{Sec.}{Secs.}
\Crefname{section}{Section}{Sections}
\Crefname{table}{Table}{Tables}
\crefname{table}{Tab.}{Tabs.}
 
\def\wacvPaperID{199} % Enter the WACV Paper ID here
\def\confName{WACV}
\def\confYear{2024}

\begin{document}

%%%%%%%%% TITLE
\title{Training-free Object Counting with Prompts}

\author{
Zenglin Shi \textsuperscript{\rm 1}, Ying Sun\textsuperscript{\rm 1, 2}, Mengmi Zhang\textsuperscript{\rm 1, 2, 3} \\
 \textsuperscript{\rm 1} \small I2R, Agency for Science, Technology and Research, Singapore \\
 \textsuperscript{\rm 2} \small CFAR, Agency for Science, Technology and Research, Singapore \\
  \textsuperscript{\rm 3} \small Nanyang Technological University (NTU), Singapore \\
}

%\and

\maketitle

\def\eg{\textit{e.g.}}
\def\ie{\textit{i.e.}}
\def\Eg{\textit{E.g.}}
\def\etal{\textit{et al. }}
\def\etc{\textit{etc.}}
\newcommand{\dimny}{\mathcal{M}\xspace}
\newcommand{\dimyH}{\mathcal{H}\xspace}
\newcommand{\dimyW}{\mathcal{W}\xspace}
\newcommand{\dimH}{\ensuremath{H}}
\newcommand{\dimW}{\ensuremath{W}}
\newcommand{\dimC}{\ensuremath{C}}
\newcommand{\nStage}{\mathcal{L}\xspace}
\newcommand{\scale}{\mathcal{S}\xspace}
\newcommand{\mypartitletwo}[2][2]{\vspace*{-#1 ex}~\\{\noindent {\bf #2}}}
\newcommand{\mypartitle}[1]{\vspace*{-3ex}~\\{\noindent \underline{\bf #1}}}
\newcommand{\todo}[1]{\textcolor{red}{\textbf{#1}}}
\newcommand{\dimn}{\ensuremath{M}}
\newcommand{\apriori}{\textit{a priori}\xspace}
\newcommand{\mapping}{\ensuremath{G}\xspace}
\newcommand{\params}{\ensuremath{\theta}\xspace}
%%% MACROS for the annotations
\newcommand{\data}{\ensuremath{X}\xspace}
\newcommand{\SV}{\ensuremath{X}\xspace}
\newcommand{\pro}{\ensuremath{P}\xspace}
\newcommand{\gt}{\ensuremath{G}\xspace}
\newcommand{\npro}{\ensuremath{N}\xspace}
\newcommand{\featSpace}{\ensuremath{\mathrm{\cal X}}\xspace}
\newcommand{\lSpace}{\ensuremath{\mathrm{\cal Y}}\xspace}
\newcommand{\labbb}{\ensuremath{\mathbf{t}}\xspace}
\newcommand{\state}{\ensuremath{z}\xspace}
\newcommand{\nframes}{\ensuremath{T}\xspace}
\newcommand{\kupdate}{\ensuremath{\boldsymbol{\varphi}}\xspace}
\newcommand{\sol}{\ensuremath{\boldsymbol{\beta}}\xspace}
\newcommand{\nsamples}{\ensuremath{N}\xspace}
\newcommand{\Msamp}{\ensuremath{M_{\mathrm{s}}}\xspace}
\newcommand{\nparticles}{\ensuremath{P}\xspace}

\newcommand{\nDepth}{\ensuremath{D_{\mathrm{max}}}\xspace}
\newcommand{\nTrees}{\ensuremath{K}\xspace}
\newcommand{\Xmat}{\ensuremath{\mathbf{X}}\xspace}
\newcommand{\Ymat}{\ensuremath{\mathbf{Y}}\xspace}
\newcommand{\HH}{\ensuremath{\mathbf{H}}\xspace}
\newcommand{\Smat}{\ensuremath{\mathbf{S}}\xspace}
\newcommand{\Dmat}{\ensuremath{\mathbf{D}}\xspace}
\newcommand{\eye}{\ensuremath{\mathbf{e}}\xspace}
\newcommand{\err}{\ensuremath{\boldsymbol{\xi}}\xspace}
\newcommand{\coeff}{\ensuremath{\mathbf{w}}\xspace}
\newcommand{\samp}{\ensuremath{\mathbf{x}}\xspace}
\newcommand{\laby}{\ensuremath{\mathbf{y}}\xspace}
\newcommand{\func}{\ensuremath{\mathbf{g}}\xspace}
\newcommand{\edv}{\mathrel\Vert}
\newcommand{\thresh}{\ensuremath{\tau}\xspace}
\newcommand{\treedepth}{\ensuremath{\Gamma_{\mathrm{depth}}}\xspace}
\newcommand{\sampler}{\emph{Sampler}}
\newcommand{\normal}{\ensuremath{\mathrm{\cal N}}\xspace}
\newcommand{\ssvmCost}{\ensuremath{\ell}\xspace}
\newcommand{\Perp}{\perp \! \! \! \perp}
\def\ci{\perp\!\!\!\perp}
\newcommand{\RR}{I\!\!R}

\newcommand{\labl}{\ensuremath{y}\xspace}

\newcommand{\mean}{\ensuremath{\mu}\xspace}

\newcommand{\Mult}{\ensuremath{\mbox{Mult}}\xspace}
\newcommand{\Cat}{\ensuremath{\mbox{Categorical}}\xspace}
\newcommand{\argmin}{\mathop{\mathrm{arg\,min}}}
\newcommand{\argmax}{\mathop{\mathrm{arg\,max}}}

\newcommand{\cs}[1]{\textcolor{red}{[\textbf{CS}: #1]}}
\newcommand{\psmm}[1]{\textcolor{orange}{[\textbf{PM}: #1]}}
\newcommand{\zl}[1]{\textcolor{blue}{[\textbf{ZL}: #1]}}
\newcommand{\sm}[1]{\textcolor{red}{[\textbf{SM}: #1]}} %Subhransu

%%%%%%%%% ABSTRACT
\begin{abstract}
    This paper tackles the problem of object counting in images. Existing approaches rely on extensive training data with point annotations for each object, making data collection labor-intensive and time-consuming. To overcome this, we propose a training-free object counter that treats the counting task as a segmentation problem. Our approach leverages the Segment Anything Model (SAM), known for its high-quality masks and zero-shot segmentation capability. However, the vanilla mask generation method of SAM lacks class-specific information in the masks, resulting in inferior counting accuracy. To overcome this limitation, we introduce a prior-guided mask generation method that incorporates three types of priors into the segmentation process, enhancing efficiency and accuracy. Additionally, we tackle the issue of counting objects specified through text by proposing a two-stage approach that combines reference object selection and prior-guided mask generation. Extensive experiments on standard datasets demonstrate the competitive performance of our training-free counter compared to learning-based approaches. This paper presents a promising solution for counting objects in various scenarios without the need for extensive data collection and counting-specific training. 
    Code is available at \url{https://github.com/shizenglin/training-free-object-counter}.
\end{abstract}
\section{Introduction}
Object counting refers to the task of estimating the number of specific objects present in an image. Traditionally, class-specific object counting approaches have been developed to count objects belonging to predefined categories such as humans, animals, or cars. These approaches, \eg, \cite{lin2001estimation, chan2008privacy,lempitsky2010learning, zhang2016single,Shi2018vlad,zhang2019nonlinear,zhang2022crossnet,rong2021coarse,yang2022crowdformer}, demonstrate excellent performance when counting objects within their trained categories. However, they face limitations in counting objects that fall outside their predefined categories during testing. On the other hand, class-agnostic object counting approaches, as seen in recent works \cite{lu2019class, ranjan2021learning, yang2021class, shi2022represent, ranjan2022exemplar, you2023few,xu2023zero}, offer a more flexible solution by enabling the counting of objects from arbitrary categories with the aid of a few support exemplars. This paper aims to contribute a training-free class-agnostic counting approach, thereby enhancing the versatility and applicability of object counting in various scenarios.

In class-agnostic counting, the dominant approach involves generating a density map through a similarity map, which compares visual features between exemplars and query images. The advantage of the similarity map is its independence from specific object classes, allowing dynamic adaptation during counting. Research efforts have focused on improving the quality of similarity maps to enhance counting accuracy \cite{gupta2021visual,zhang2021look,zhang2018finding,lu2019class, ranjan2021learning, yang2021class, shi2022represent, you2023few}. Once a high-quality similarity map is obtained, the goal is to learn a model that maps it to the corresponding density map. The count is derived by summing the density values. However, these methods typically require a large amount of training data with point annotations for each object, making data collection labor-intensive and time-consuming. As a result, scaling density-based counting approaches across multiple visual categories becomes challenging.
\begin{figure}[t!]
\centering
\includegraphics[width=0.99\linewidth]{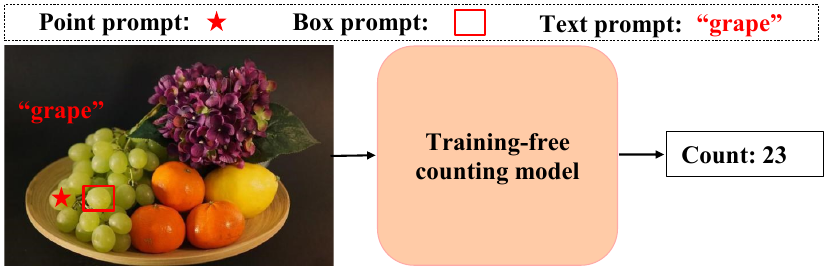}
\caption{\textbf{The objective of this work} is to build a training-free object counting model, where we can specify what to count with prompts such as points, boxes, or texts.
}
%\vspace{-4mm}
\label{fig:fig1} 
\end{figure}

To address the challenges mentioned earlier, this paper aims to develop a training-free object counter capable of counting specified objects through input prompts such as points, boxes, or texts, as illustrated in Fig. \ref{fig:fig1}.
To achieve this, the counting task is formulated as a segmentation problem. Specifically, a segmentation model is employed to identify and separate individual objects specified by input prompts, resulting in a set of binary segmentation maps corresponding to different target objects. The estimated object count is then obtained by counting the number of these maps. For the segmentation model, the Segment Anything Model (SAM) is considered due to its exceptional ability to generate high-quality masks and perform zero-shot segmentation in diverse scenarios using input prompts like points or boxes. However, the vanilla mask generation method of SAM alone does not produce satisfactory results, as it lacks class-specific information in the masks. To mitigate this issue, a prior-guided mask generation method is introduced by incorporating three types of priors into the segmentation process of SAM. These priors serve as additional guidance to refine the mask generation process and enhance counting accuracy. Furthermore, we tackle the challenge of counting objects specified through text. To address this, we propose a two-stage approach that combines reference object selection and prior-guided mask generation. This approach enables accurate counting of objects specified through textual prompts.

Overall, we make three contributions in this paper: (\textit{i}) We approach the class-agnostic counting task as a prompt-based segmentation problem. By doing so, we eliminate the need for extensive data collection and model training, thereby making counting more accessible to the public.
 (\textit{ii}) We propose a new prior-guided mask generation method that improves the efficiency and accuracy of the segmentation process in SAM by incorporating three types of priors. 
 (\textit{iii}) We present a new two-stage approach for counting objects specified through text, combining reference object selection with the prior-guided mask generation method. Through extensive experiments on standard datasets, we validate the competitive performance of our training-free counter when compared to learning-based approaches.

\section{Related Works}
\subsection{Learning-based object counting}
Class-specific object counting focuses on counting objects belonging to predefined categories, like humans, animals, or cars. The dominant approach is to employ regression-based methods to generate density maps. This method, initially proposed by Lempitsky et al. \cite{lempitsky2010learning}, has been the foundation for subsequent works \cite{zhang2016single,Shi_2018_CVPR,shi2019counting,cheng2019learning,cheng2022rethinking,cheng2019improving,yang2022crowdformer,lian2019density,lian2021locating,xu2021crowd}. Density-based counting requires point annotations for each countable object in training images. These points are convolved with Gaussian kernels to create density maps for training. A model is then trained to predict a density map for each input image, and the object count is obtained by summing the pixel values in the predicted density map. While class-specific counters perform well on trained categories, they lack flexibility in counting objects outside their predefined categories during testing.

Class-agnostic object counting aims to count objects of arbitrary categories using only a few support exemplars \cite{lu2019class, zhang2019nonlinear, ranjan2021learning, yang2021class, shi2022represent, you2023few,gupta2021visual,zhang2021look,zhang2018finding,ranjan2022exemplar,xu2023zero,shi2023focus}. Unlike class-specific counting relying on predefined common objects in training images, class-agnostic counting allows users to define and customize objects of interest using support exemplars. While density map prediction remains prevalent, class-agnostic methods learn a mapping from similarity maps to density maps. The key is that the similarity map is independent of specific object classes, enabling dynamic adaptation for counting arbitrary classes. Research focuses on improving similarity map quality \cite{lu2019class, ranjan2021learning, yang2021class, shi2022represent, you2023few} and addressing issues like test-time adaptation \cite{ranjan2022exemplar} and the need for human-annotated exemplars \cite{ranjan2022exemplar,xu2023zero}.

Density-based counting methods often demand extensive training data with point annotations for each object, which can be a laborious process when dealing with millions of objects across thousands of images. Scaling these methods across various visual categories becomes challenging. To overcome this, we introduce a training-free object counter. It allows object counting using prompts like points, boxes, or text, eliminating the need for training. This approach broadens the possibilities for counting objects in diverse scenarios, without the data collection and counting-specific training burden

\subsection{Prompt-based foundation model}

The emergence of large language models like ChatGPT has transformed the field of natural language processing and extended to computer vision. These ``foundation models" exhibit impressive generalization in zero-shot and few-shot scenarios. In computer vision, CLIP is a notable foundation model that utilizes contrastive learning to train text and image encoders, enabling it to handle novel visual concepts and data distributions through text prompts \cite{radford2021learning}. CLIP demonstrates exceptional zero-shot transfer capabilities across diverse visual domains. Another foundation model, the segment anything model (SAM), is designed for image segmentation \cite{kirillov2023segment}. SAM utilizes prompts like points and boxes to generate high-quality object masks, achieving remarkable performance on various segmentation benchmarks and showcasing zero-shot transfer abilities across diverse datasets. 

\begin{figure*}[!t]
\centering
\begin{subfigure}{0.9\linewidth}
\includegraphics[width=\textwidth]{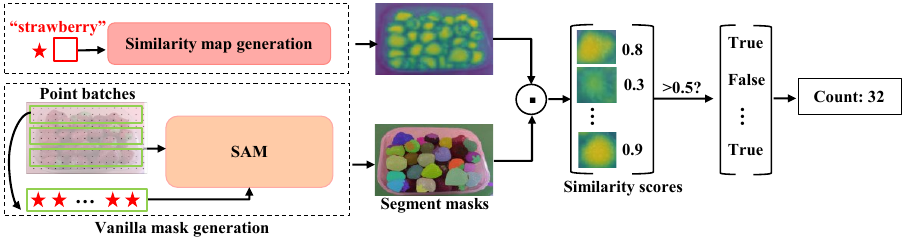}
\caption{\textbf{Vanilla training-free counting model}}
\label{fig:model-a}    
\end{subfigure}
\begin{subfigure}{0.8\linewidth}
\includegraphics[width=\textwidth]{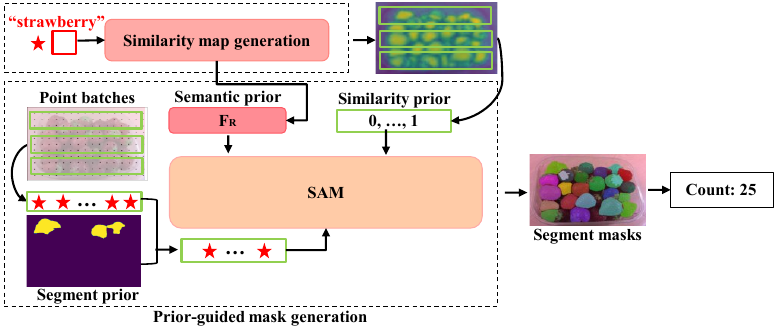}
\caption{\textbf{Prior-guided training-free counting model}}
\label{fig:model-b}    
\end{subfigure}
\caption{\textbf{Training-free counting model with prompts.} \textbf{(a)}: The first step in the vanilla method is to create a similarity map from the input prompts (see Figure \ref{fig:sim}). Then, we generate masks for all objects in the image using point grids as prompts, which are processed in batches. To score each object indicated by a mask, we calculate the average similarity within the masked area. If the score exceeds a preset threshold, e.g., $0.5$, we identify it as a target object. Finally, we count all the identified target objects to determine the total.
\textbf{(b)}: Our advanced method improves SAM's mask generation by integrating three key priors. Firstly, we create a similarity map using input prompts, as in the vanilla method, which helps label positive and negative points. Secondly, we maintain an overall segment map that includes all current segmented regions, avoiding redundant processing by checking for existing points. Thirdly, we use the reference object feature $F_R$ as a semantic prior, enhancing SAM's ability to identify and segment target objects in the image. These priors enable us to focus solely on target objects, improving segmentation accuracy. The object count is determined by the number of output segment maps.
}
\label{fig:model}    
\end{figure*}
These foundation models have revolutionized the field of computer vision, offering exciting possibilities for powerful generalization and the capacity to tackle novel tasks and data distributions without requiring explicit training on those specific instances. Building upon the capabilities of CLIP and SAM, we propose a training-free object counter in this work, pushing the boundaries of zero-shot learning and generalization capabilities in the field of object counting.

\section{Training-free Counting Method}
This paper considers the problem of class-agnostic counting, in which the category of objects is specified by input prompts such as points, bounding boxes, or texts. To tackle this problem, we present a new approach that eliminates the need for training. Our method leverages the segmentation foundation model as its basis, enabling accurate and efficient counting. A comprehensive overview of our method is provided in Fig.~\ref{fig:model}.

\subsection{Counting by segmentation}
\label{sec:countSeg}
We formulate our counting task as a segmentation problem. Specifically, we employ a segmentation model to identify and separate the individual objects specified by input prompts from an image, resulting in a set of binary segmentation maps corresponding to different target objects. The estimated object count is calculated as the number of these maps.

The segmentation model $f$ plays a crucial role in determining the accuracy of object counting. In this paper, we leverage SAM \cite{kirillov2023segment} as the foundation model for segmentation. SAM has been shown to generate high-quality masks and perform zero-shot segmentation in diverse scenarios, using input prompts such as points or boxes. SAM comprises three essential components: an image encoder $f_{ie}$, a prompt encoder $f_{pe}$, and a mask decoder $f_{md}$. During the segmentation process, SAM first utilizes $f_{ie}$ to extract the features of the input image, while $f_{pe}$ encodes the provided prompts. The encoded image and prompts are then fed into $f_{md}$, which produces the final mask output. 

To incorporate SAM into our counting task, we start by presenting a vanilla method, as illustrated in Fig. \ref{fig:model-a}. In particular, given an image $I$ and input prompts $P$, we first generate binary segment masks of the reference objects indicated by the input prompts, denoted as $S{=}f(I, P)$. Meanwhile, we can obtain the image feature $F_I{=}f_{ie}(I)$ and compute the feature of the reference objects by element-wise multiplication between the reference masks and the image feature, denoted as $F_R {=}F_I\odot S$ where $\odot$ denotes the Hadamard product. The similarity map $Sim$ between the image feature $F_I$ and the reference feature $F_R^{(k)}$ is computed with the cosine similarity metric. Subsequently, we proceed to generate masks for all objects present in the image by employing point grids (32 points on each side) as prompts to segment the entire scene, resulting in $\{mask^{(1)},\dots, mask^{(m)}\}$. The similarity score of each object indicated by each mask is calculated by averaging the masked similarity values, denoted as $score^{(i)} {=} \overline{Sim\odot mask^{(i)}}$. If the $score^{(i)}$ surpasses a predefined threshold $\epsilon$, we consider $mask^{(i)}$ as indicating a target object. Finally, we determine the total count by tallying all the identified target objects. 

Despite the impressive performance of the vanilla method in various scenarios, we have identified two limitations that hinder its overall efficiency and accuracy. Firstly, the post-processing step, which involves determining target objects from all objects in the image using a similarity map, is not as efficient as desired. This process requires segmenting all objects, which can be computationally expensive and time-consuming. Secondly, determining the appropriate similarity score threshold $\epsilon$ presents a significant challenge. The similarity map is not flawless, making it difficult to select an optimal threshold. A high threshold may lead to missed target objects, resulting in an underestimation of the object count, while a low threshold may count non-target objects, leading to an overestimation of the count. To address these limitations, we propose a new approach called prior-guided automatic mask generation. This approach leverages prior knowledge to improve the efficiency and accuracy of object counting.  
\begin{figure}[t!]
\centering
\includegraphics[width=0.99\linewidth]{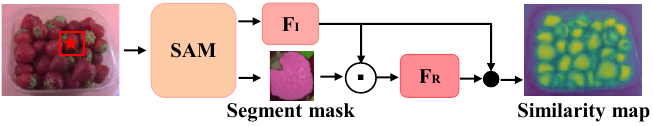}
\caption{\textbf{Similarity map generation with point or box prompt.} To process the input image and prompt, we employ SAM, which allows us to extract the image feature $F_I$ and obtain the reference object mask. By multiplying the image feature with the reference object mask, we obtain the reference object feature $F_R$. The similarity between the image feature $F_I$ and the reference feature $F_R$ is calculated using cosine similarity. $\odot$ and $\bullet$ denote the Hadamard product and the Euclidean dot product, respectively.
}
\label{fig:sim}    
\end{figure}

\subsection{Prior-guided mask generation}
\label{sec:prior}
In order to generate masks for all target objects in the image, it would be ideal to have specific prompts for each target object. However, in practice, we often only have prompts for a few target objects. To overcome this limitation, we employ a regular grid of $t \times t$ points that cover the entire image as prompts to generate masks. However, this approach may segment non-target objects, leading to inaccurate results. To address this challenge, we incorporate three types of priors into the segmentation process. These priors assist in distinguishing between positive and negative points within the grid, ensuring that only the desired target objects are accurately segmented. A visual illustration of our method is provided in Fig.~\ref{fig:model-b}.

\textbf{Similarity prior.} To incorporate the similarity prior, we compute the cosine similarity between the reference object feature and the image feature using the cosine similarity metric, resulting in a similarity map, as illustrated in Fig. \ref{fig:sim}. We then apply Otsu's binarization approach \cite{otsu1979threshold} to create a binary similarity map, serving as the label map.
In the label map, points in the grids corresponding to a value of $1$ are considered positive points, while the rest are regarded as negative points. By utilizing this label map, SAM can effectively focus on segmenting the contiguous regions surrounding the positive points while disregarding the negative ones. This incorporation of the similarity prior enhances the segmentation process by leveraging the information from the similarity map, leading to more precise identification and separation of the target objects within the image.
\begin{figure*}[t!]
\centering
\includegraphics[width=0.99\linewidth]{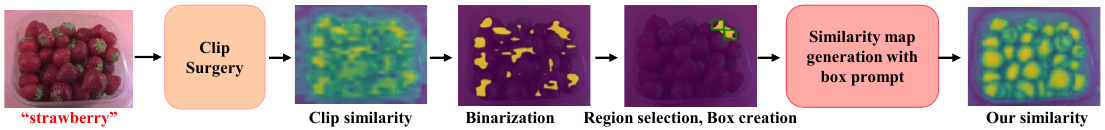}
\caption{\textbf{Similarity map generation with a text prompt.} Our approach starts with a coarse similarity map computed using CLIP-Surgery. From this map, we select reference objects through several steps. First, we binarize the similarity map. Then, we identify the largest connected component, likely containing the target objects. Further refinement is achieved by creating bounding boxes around sub-regions within this component. These bounding boxes serve as prompts for SAM, resulting in a high-quality similarity map (see Fig. \ref{fig:sim}).
}
\label{fig:clip}    
\end{figure*}

\textbf{Segment prior.} To address the computational memory constraints of SAM when processing all points from the grids simultaneously, the points are divided into batches for sequential processing. In this way, we can leverage the segments generated from the first batches as priors to guide the segmentation of the subsequent batches. Since multiple points from grids may be used as prompts for an object, these points may result in redundant processing by SAM, leading to the generation of multiple masks for the same object. This redundancy is both inefficient and inaccurate. To address this issue, we maintain an overall segment map that contains all the segmented regions up to the current batch. When processing the current batch, if any points from the batch are already present in the segment regions, we remove those points from the current batch. Also, if there are no positive points in the updated batch, this batch won't be processed by SAM. By leveraging the overall segment map, SAM can refine and adjust its segmentation predictions based on the already segmented regions. This approach enhances the efficiency and accuracy of the segmentation process by avoiding redundant computations and ensuring consistency across different batches.

\textbf{Semantic prior.} In addition to the similarity prior and segment prior, which aid in selecting positive and distinctive points as prompts for segmenting target objects, we recognize that the point prompts alone may not provide sufficient information to accurately segment these objects. To address this limitation, we propose incorporating the reference object feature $F_R$ as a semantic prior, enabling the mask decoder $f_m$ to better identify and segment the target objects within the image. By integrating the reference object feature $F_R$, the modified mask generation process becomes $S = f_{md}(F_I, F_R, f_{pe}(P))$ instead of $S = f_{md}(F_I, f_p(d))$, where $P$ represents the point prompts. Notably, the reference object feature $F_R$ is extracted from the well-trained image encoder $f_{ie}$ and does not require any fine-tuning. By incorporating the reference object feature $F_R$ and modifying the mask generation process as mentioned, the mask decoder can now focus more attentively on the reference object feature. This inclusion of additional contextual information through $F_R$ greatly assists the mask decoder in accurately distinguishing and segmenting the target objects within the image. 

By utilizing these priors, we can focus exclusively on the target objects without segmenting unrelated entities, resulting in improved efficiency and accuracy compared to the vanilla method.

\subsection{Text-specified mask generation}
In addition to the previous approaches, we also address the challenge of counting objects specified through text, where the object of interest is described using textual information rather than explicit points or bounding boxes. To tackle this problem, we present a two-stage approach that combines reference object selection and prior-guided mask generation.  

\textbf{Reference object selection.} In the first stage, we leverage the CLIP-Surgery \cite{li2023clip}, an enhanced version of CLIP \cite{radford2021learning}, to compute the similarity between the image and text representations at the pixel level. The improved architecture and feature extraction of CLIP-Surgery enables more accurate computation of image-text similarity. However, the initial similarity map obtained from CLIP-Surgery may not be of high quality, making it unsuitable for direct use in mask generation. Therefore, we leverage the initial similarity map to select reference objects.

To facilitate accurate identification of the target objects, we employ Otsu's binarization approach \cite{otsu1979threshold} on the initial similarity map, resulting in a binary similarity map. This binary map acts as a guide for selecting the region that is most likely to contain the target objects. From the binary similarity map, we extract the largest connected component, which represents the primary region that encompasses the target objects. To ensure precise localization, we further divide the contour of the largest connected component into multiple sub-contours. For each sub-contour, we create a corresponding bounding box. This strategy helps us avoid including irrelevant objects within the bounding boxes and focuses solely on the target objects. To handle potential overlaps among the resulting bounding boxes, we apply the Non-Maximum Suppression (NMS). NMS enables us to select the most appropriate and non-overlapping bounding boxes, ensuring that we retain only the most accurate representations of the target objects. 

\begin{table*}[t!]
\centering
\resizebox{1.7\columnwidth}{!}{
\begin{tabular}{@{}lccccccccccccc@{}}
\toprule
&\multirow{2}{*}{\textbf{Training}} & \multirow{2}{*}{\textbf{Prompt}} & \multicolumn{4}{c}{\textbf{FSC-147}} & \multicolumn{4}{c}{\textbf{CARPK}}\\
\cmidrule(lr){4-7} \cmidrule(lr){8-11}
& & & \textbf{MAE} $\downarrow$  &
\textbf{RMSE} $\downarrow$ & \textbf{NAE} $\downarrow$ &
\textbf{SRE} $\downarrow$ &\textbf{MAE} $\downarrow$  &
\textbf{RMSE} $\downarrow$ & \textbf{NAE} $\downarrow$ &
\textbf{SRE} $\downarrow$  \\
\hline
GMN \cite{lu2019class} &Yes&box&26.52 &124.57 &-&-7.48 &9.90 &- &-\\
FamNet+ \cite{ranjan2021learning} &Yes&box&22.08 &99.54 &0.44&6.45&18.19 &33.66 &- &-\\
CFOCNet+ \cite{yang2021class} &Yes&box&22.10 &112.71 &-&-&-&-&-&-\\
BMNet+ \cite{shi2022represent} &Yes&box&\underline{14.62} &\underline{91.83} &\underline{0.25}&\underline{2.74}&\underline{5.76} &\underline{7.83} &- &-\\
SAM &No&N.A.&42.48 &137.50 &1.14&8.13&16.97 &20.57 &0.70 &5.30\\
Ours (vanilla) &No&box&26.29 &137.89 &0.38&4.38&15.67 &19.44 &0.67 &5.06\\
\rowcolor{Gray}
Ours &No&box&\textbf{19.95} &\textbf{132.16} &\textbf{0.29}&\textbf{3.80}&\textbf{10.97}&\textbf{14.24} &\textbf{0.48} &\textbf{3.70}\\
\hline
Ours (vanilla) &No&point&25.18 &137.62 &0.37 &4.34&15.67 &19.44 &0.67 &5.06\\
\rowcolor{Gray}
Ours &No&point&20.10 &132.83 &0.30 &3.87&11.01 &14.34 &0.51 &3.89\\

\bottomrule
\end{tabular}}
\caption{\textbf{Effect of our approaches with point and box prompt on FSC-147.} Our methods demonstrate competitive performance compared to learning-based approaches. The \textbf{bold} font highlights the best counting results among training-free methods, while the \underline{underlined} font indicates the best counting results among learning-based methods. This convention is consistent throughout the tables presented below.
}
\label{tab:box-prompt}
\end{table*}

\textbf{Prior-guided mask generation.} In the second stage of our approach, we utilize the bounding boxes obtained from the first stage as prompts for SAM. SAM generates masks corresponding to the reference objects identified in the previous stage. Once we have obtained the masks of the reference objects, we proceed to compute a similarity map between the reference feature and the image feature using the cosine similarity measure, as illustrated in Figure \ref{fig:clip}. This similarity map captures the resemblance between the reference objects and the image regions. To generate accurate masks for all target objects, we employ our prior-guided mask generation approach, as detailed in Section \ref{sec:prior}, which leverages the information from the similarity map.

\section{Experiments}
\subsection{Experimental setup}
\textbf{Datasets.} We evaluate our approaches on two commonly used counting datasets, namely, \textbf{FSC147} \cite{ranjan2021learning} and \textbf{CARPK} \cite{hsieh2017drone}. The FSC147 dataset comprises 6135 images from 147 distinct object categories. We utilize the testing set, which includes 1190 images from 29 object categories, for evaluation as our approach does not require training. In the CARPK dataset, there are 1448 images with approximately 90,000 cars captured from a drone view. The testing set consists of 459 images.

\textbf{Evaluation metrics.} We report the Mean Absolute Error (MAE), Root Mean Square Error (RMSE), Normalized Relative Error (NAE) and Squared Relative Error (SRE) metrics given count estimates $\hat{y}$ and their ground-truth $y$ for $n$ test images. In particular, MAE = $\frac{1}{n} \sum_{i=1}^n |y_i- \hat{y}_i|$, RMSE = $\sqrt{\frac{1}{n} \sum_{i=1}^n (y_i- \hat{y}_i)^2}$, NAE = $\frac{1}{n} \sum_{i=1}^n {|y_i- \hat{y}_i|/y_i}$, SRE = $\sqrt{\frac{1}{n} \sum_{i=1}^n {(y_i- \hat{y}_i)^2/y_i}}$.

\textbf{Implementation details.} For our experiments, we utilize the ``vit\_b" image encoder in SAM. To incorporate box prompts, we use annotated reference object boxes provided by the dataset and extract their center points as point prompts. When using text prompts, we utilize the template "the photo of many" followed by the specific object class name derived from the dataset. In our vanilla counting method (Section \ref{sec:countSeg}), setting the similarity score threshold $\epsilon$ to 0.5 yields the best performance across datasets. For our prior-guided mask generation method (Section \ref{sec:prior}), we use a regular grid of $t \times t$ points as prompts for SAM. The default value of $t$ in the original SAM paper is 32, but we observed that a larger $t$ is more effective when counting small objects. Thus, we propose setting $t$ dynamically as $t {=} (32 \edv O_{size}+1)*32$, where $O_{size}$ represents the minimum size of reference objects obtained from their masks, and $\edv$ denotes exact division.

\subsection{Counting with point and box prompts}
\textbf{Baseline methods.} In this experiment, we evaluate our counting approach utilizing both point and box prompts. To assess its effectiveness, we compare it against four other learning-based class-agnostic counting approaches that employ diverse methods for learning high-quality similarity maps. The compared approaches include GMN \cite{lu2019class}, FamNet+ \cite{ranjan2021learning}, CFOCNet+ \cite{yang2021class}, and BMNet+ \cite{shi2022represent}. It is important to note that these learning-based approaches are specifically designed to accept only boxes as prompts. In addition to the aforementioned comparisons, we also report the counting results obtained through automatic map generation with SAM, denoted as SAM. This allows us to evaluate the performance of SAM's default behavior in counting objects without any further processing.

\textbf{Results.}
As shown in Table~\ref{tab:box-prompt}, direct counting using SAM yields the poorest performance due to the lack of class information in the mask generated by SAM. This leads to the inclusion of numerous non-target objects during counting. However, by utilizing the similarity map for target object selection, our vanilla method improves performance significantly. For instance, on the FSC-147 dataset with box prompts, MAE decreases from 42.48 to 26.29. Furthermore, our method incorporates three priors to enhance differentiation between target and non-target objects during segmentation, resulting in a further reduction of MAE from 26.29 to 19.95. When compared to four learning-based methods, our approach outperforms three of them. The best learning-based method, BMNet+ \cite{shi2022represent}, achieves a mere 5.33 reduction in MAE and a 0.04 reduction in NAE on the FSC-147 dataset, as well as a 5.21 reduction in MAE on the CARPK dataset, despite using thousands of training data and complex model design and training procedures. Our methods still retain their advantages when using point prompts. These results demonstrate the promising nature of our training-free method.

Fig.~\ref{fig:case-a} demonstrates the effectiveness of our methods through success and failure results. Our approach consistently achieves accurate counting, even in challenging scenes with sparse small objects or dense large objects (first row). However, in extreme scenes where individual objects are too tiny to be distinguishable or objects blend with the background (last row), our approach encounters difficulties. These scenarios present ongoing challenges in counting tasks.
\begin{table}[!t]
\centering
\resizebox{0.95\columnwidth}{!}{
\begin{tabular}{@{}lccccccccccccc@{}}
\toprule
&\multirow{2}{*}{\textbf{Training}} & \multicolumn{4}{c}{\textbf{FSC-147}}\\
\cmidrule(lr){3-6}
& & \textbf{MAE} $\downarrow$ &
\textbf{RMSE} $\downarrow$ & \textbf{NAE} $\downarrow$ &
\textbf{SRE} $\downarrow$ \\
\hline
Xu \etal \cite{xu2023zero} &Yes&\underline{22.09} &\underline{115.17} &\underline{0.34}&\underline{3.74}\\
SAM &No&42.48 &137.50 &1.14&8.13\\
Ours (vanilla) &No&32.86 &142.89 &0.44 &5.12&\\
\rowcolor{Gray}
Ours &No&\textbf{24.79} &\textbf{137.15} &\textbf{0.37}&\textbf{4.52}\\

\bottomrule
\end{tabular}}
\caption{\textbf{Effect of our approaches with text prompt on FSC-147.}
Our methods yield competitive counting results compared to the learning-based method.
}
\label{tab:text-prompt}
\end{table}
\subsection{Counting with text prompt}
\textbf{Baseline methods.} In this experiment, we evaluate our counting approach using text prompts. To assess its effectiveness, we compare it against the learning-based method proposed by Xu et al. \cite{xu2023zero}, which also utilizes text to specify the target objects. We again report the counting results obtained from SAM alone for comparison.  

\textbf{Results.} Table \ref{tab:text-prompt} presents the results. Our vanilla method significantly improves performance by utilizing the similarity map obtained with Clip-Surgery to select target objects compared to direct counting with SAM alone. For instance, on the FSC-147 dataset, the MAE decreases from 42.48 to 32.86. To further enhance the quality of the similarity map obtained through Clip-Surgery, we introduce our reference object selection algorithm. Combining this with our prior-guided mask generation method yields outstanding counting results, reducing MAE from 32.86 to 24.79. Notably, the learning-based approach proposed by Xu et al. \cite{xu2023zero} only slightly outperforms our method, with a difference of 2.7 in MAE and a marginal 0.03 in NAE.

\begin{figure*}[t!]
\centering
\begin{subfigure}{0.99\linewidth}
\includegraphics[width=\textwidth]{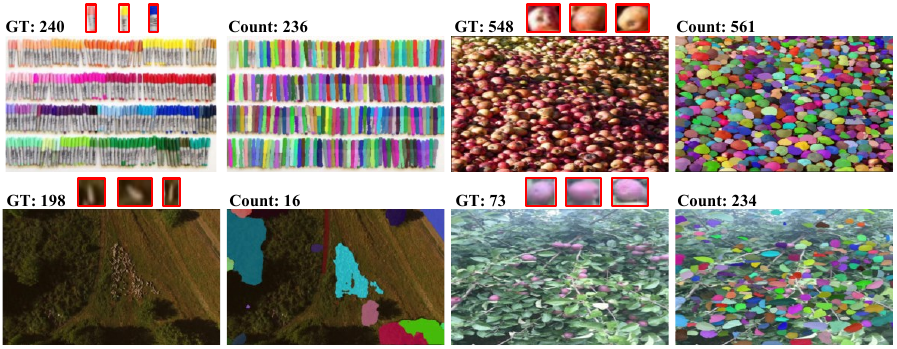}
\caption{\textbf{Success (first row) and failure (last row) results with box prompt}}
\label{fig:case-a}    
\end{subfigure}
\begin{subfigure}{0.99\linewidth}
\includegraphics[width=\textwidth]{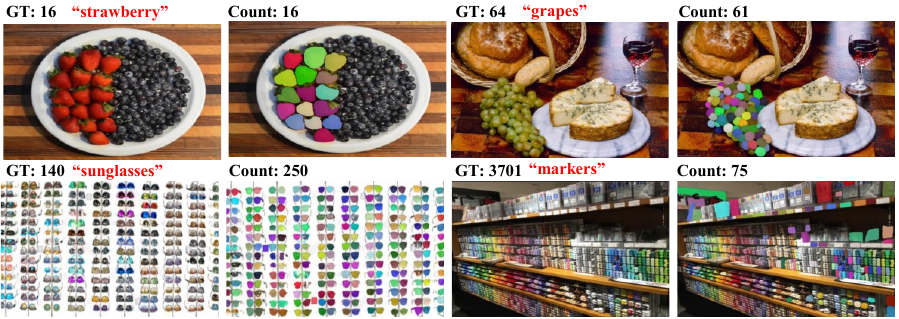}
\caption{\textbf{Success (first row) and failure (last row) with text prompt}}
\label{fig:case-b}    
\end{subfigure}
\caption{\textbf{Success and failure results with box prompts (Row 1 and 2) and with text prompts (Row 3 and 4)}. When objects are individually visible, our approaches can count them accurately. Further improvements are required in extreme scenes where individual objects are hard to distinguish, or where objects blend with the background.
}
\label{fig:case}    
\end{figure*}

Fig.~\ref{fig:case-b} illustrates the effectiveness of our methods through success and failure results. Our approach demonstrates accurate counting of target objects, even in the presence of complex backgrounds (first row). However, it may encounter double counting by segmenting each similar component. For instance, in the second row, our approach might count each lens of a pair of sunglasses as separate objects rather than recognizing them as a single entity. Additionally, counting extremely dense objects poses a challenge for our method.

\subsection{Ablation study and analysis}
\begin{table}[!t]
\centering
\resizebox{0.99\linewidth}{!}{
\begin{tabular}{ccccccccc}
\toprule
\multicolumn{3}{c}{\textbf{Three types of priors}} & \multicolumn{4}{c}{\textbf{FSC-147}}\\
\cmidrule(lr){1-3} \cmidrule(lr){4-7}
\textbf{Similarity} & \textbf{Segment} & \textbf{Semantic} &\textbf{MAE} $\downarrow$ & \textbf{RMSE} $\downarrow$ &\textbf{NAE} $\downarrow$ & \textbf{SRE} $\downarrow$  \\
\hline
 & & &42.48 &137.59& 1.14 & 8.13 \\
 \checkmark & & &21.36 &134.07&\textbf{0.27}&4.29\\
 &\checkmark &  &26.14 &134.98&0.51 &4.84\\
 & & \checkmark&37.17 &134.86&1.12 &8.19\\
 \checkmark&\checkmark &  &20.38 &134.32&0.31 &3.89\\
 \checkmark&& \checkmark  &20.83 &133.16&0.38 &5.29\\
%&\checkmark & \checkmark &34.43 &135.06&0.90 &6.65\\
\rowcolor{Gray}
\checkmark&\checkmark & \checkmark &\textbf{19.95} & \textbf{132.16} & 0.29 & \textbf{3.80} \\
\bottomrule
\end{tabular}
}
\caption{\textbf{Effect of three priors of our prior-guided mask generation method} on FSC-147 with box prompt. Each prior matters for improving the counting performance.
}
\label{tab:prior}
\end{table}
\textbf{Prior-guided mask generation.}
Our enhanced mask generation method, guided by prior information, integrates three types of priors into SAM's segmentation process, thereby boosting the counting performance. This study aims to evaluate the significance of each prior. Table \ref{tab:prior} presents the results of our experiment on the FSC-147 dataset. It is evident that each prior plays a crucial role in improving the counting performance compared to relying solely on SAM for direct counting. Moreover, when we combine the similarity prior with either the segment prior or the semantic prior, we observe additional improvements. The best counting outcomes are obtained when all three priors are integrated.

\textbf{Reference object selection.} 
We aim to evaluate the significance of our reference object selection algorithm in enhancing text-specified counting. We compare our algorithm to a baseline method that does not utilize the algorithm to enhance the quality of the similarity map obtained through Clip-Surgery. Essentially, the baseline method directly employs the similarity map obtained from Clip-Surgery for prior-guided mask generation. The results, as shown in Table \ref{tab:ros}, clearly demonstrate the improved counting performance achieved by our approach.
\begin{table}[!h]
\small
\centering
\resizebox{0.95\linewidth}{!}{
\begin{tabular}{lcccc}
\toprule
Reference object selection &\textbf{MAE} $\downarrow$ &
\textbf{RMSE} $\downarrow$ & \textbf{NAE} $\downarrow$ &
\textbf{SRE} $\downarrow$    \\
\hline
 & 39.97 &147.58 &0.47 &5.52   \\
\rowcolor{Gray} 
\checkmark &\textbf{24.79} &\textbf{137.15} &\textbf{0.37}&\textbf{4.52} \\
\bottomrule
\end{tabular}
}
\caption{\textbf{The effect of our reference object selection algorithm} on FSC-147 with text prompt. Our algorithm is crucial in improving counting accuracy. }
\label{tab:ros}
\end{table}

\textbf{Speed analysis.} Finally, we analyze the speed of our method. All tests are conducted on a machine equipped with an Nvidia RTX A5000 GPU. In GPU mode, our method achieves a processing time of 2.1 seconds per image. This speed surpasses both the 4.7 seconds of our vanilla method and the 3.4 seconds of SAM.

\section{Conclusions and future work}
This paper addresses the problem of object counting in images by introducing a training-free counter that leverages the segmentation foundation model, SAM. We propose a new prior-guided mask generation method to enhance the segmentation process in SAM by incorporating three types of priors including similarity prior, segment prior, and semantic prior. Through extensive ablation studies, we demonstrate the significant impact of each prior and their combined effect on improving counting efficiency and accuracy. Moreover, we address the problem of counting objects specified through text by presenting a two-stage approach. This approach combines reference object selection and prior-guided mask generation.  We have shown that reference object selection plays a crucial role in refining the similarity maps, enabling accurate object counting based on textual descriptions. Extensive experiments performed on standard datasets demonstrate the competitive performance of our training-free object counter when compared to learning-based approaches. 

In our future work, we aim to enhance our methods for counting occluded and tiny objects. This will involve developing more advanced adaptive thresholding methods or fine-tuning SAM using a few annotated data.

\section{Acknowledgment}
This research is supported by the National Research Foundation, Singapore under its AI Singapore Programme (AISG Award No: AISG2-RP-2021-025), its NRFF award NRF-NRFF15-2023-0001. We also acknowledge Mengmi Zhang's Startup Grant from Agency for Science, Technology, and Research (A*STAR), and Early Career Investigatorship from Center for Frontier AI Research (CFAR), A*STAR.

{\small
\bibliographystyle{ieee_fullname}
\bibliography{egbib}
}

\end{document}